\documentclass[conference]{IEEEtran}
\IEEEoverridecommandlockouts
\usepackage{cite}
\usepackage{amsmath,amssymb,amsfonts}
\usepackage{algorithmic}
\usepackage{balance}
\usepackage{bm}
\usepackage{graphicx}
\usepackage{epsfig}
\usepackage{epstopdf}
\usepackage{subfigure} 
\usepackage{textcomp}
\usepackage{mathrsfs}
\usepackage{xcolor}
\usepackage{multirow}
\usepackage{url}

\def\BibTeX{{\rm B\kern-.05em{\sc i\kern-.025em b}\kern-.08em
    T\kern-.1667em\lower.7ex\hbox{E}\kern-.125emX}}
\begin{document}
\title{A Robust Roll Angle Estimation Algorithm\\ Based on Gradient Descent}
\author{Rui Fan$^1$,~\IEEEmembership{Member,~IEEE}, Lujia Wang$^2$,~\IEEEmembership{Member,~IEEE}, \\Ming Liu$^1$,~\IEEEmembership{Senior Member,~IEEE}, Ioannis Pitas$^3$,~\IEEEmembership{Fellow,~IEEE},
	\\$^1$\textit{Robotics and Multi-Perception Laboratory, Robotics Institute, }
	\\\textit{the Hong Kong University of Science and Technology, Hong Kong. }
	\\$^2$\textit{Shenzhen Institutes of Advanced Technology, Chinese Academy of Sciences, Shenzhen, China.}
	\\$^3$\textit{School of Informatics, Aristotle University of Thessaloniki, Thessaloniki, Greece. }
	\\\texttt{\{eeruifan, eelium\}@ust.hk},\ \texttt{lj.wang1@siat.ac.cn}, \ \texttt{pitas@aiia.csd.auth.gr}
}
\maketitle

\begin{abstract}
This paper presents a robust roll angle estimation algorithm, which is developed from our previously published work, where the roll angle was estimated from a dense disparity map by minimizing a global energy using golden section search algorithm. In this paper, to achieve greater computational efficiency, we utilize gradient descent to optimize the aforementioned global energy. The experimental results illustrate that the proposed roll angle estimation algorithm takes fewer iterations to achieve the same precision as the previous method. 
\end{abstract}

\section{Introduction}
\label{sec.introduction}
Stereo vision is one of the most commonly used technologies for 3D information extrapolation \cite{Fan2018e, Trucco1998, Fan2018f}. This technology extrapolates the depth of the real-world scenery by comparing the difference between a pair of stereo images \cite{Qian1997, Fan2017a}. The relative positional difference between each pair of correspondence points is generally known as disparity \cite{Trucco1998, Fan2018g}. The disparity information enables autonomous systems to segment road regions more effectively  \cite{Fan2019}. For example, Labayrade et al. \cite{Labayrade2002} introduced an accurate obstacle detection method based on the analysis of a so-called ``v-disparity'' image, which was created by computing the histogram of each horizontal row of the disparity map. Since then, many algorithms based on v-disparity image analysis have been proposed for road region segmentation \cite{Soquet2007, Fan2018d, Yiruo2013}. These algorithms generally hypothesize that the road surface is planar and the stereo rig baseline is perfectly parallel to the horizontal road surface \cite{Broggi2005, hu2005complete}. These assumptions make it feasible to represent the road disparity projections using a linear model \cite{Chen2018, Oniga2015, Song2017}. The road regions can, therefore, be extracted by comparing the difference between the actual disparity values and the estimated linear road disparity projection model \cite{Ma2018}.

\begin{figure}[!t]
	\begin{center}
		\centering
		\includegraphics[width=0.25\textwidth]{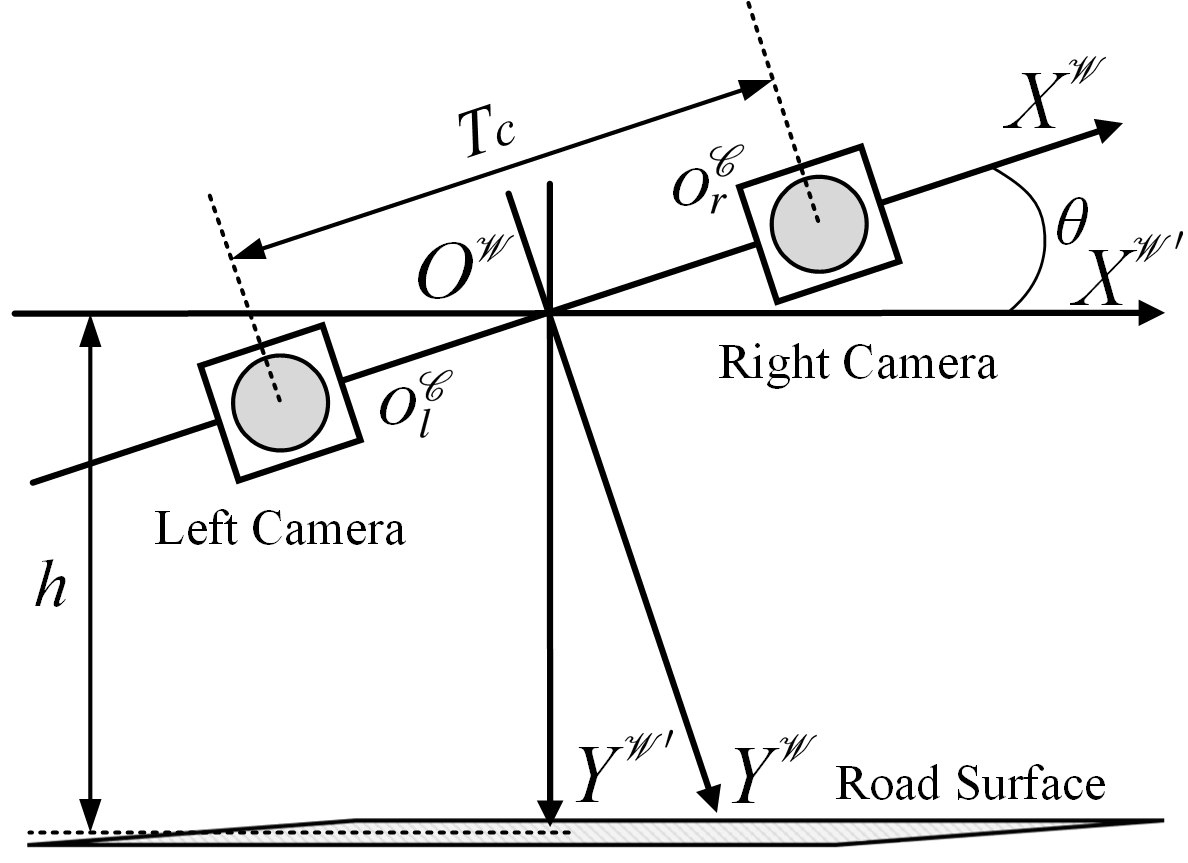}
		\centering
		\caption{Roll angle definition in a stereo rig. }
		\label{fig.roll_angle}
	\end{center}
\end{figure}

\begin{figure}[!t]
	\begin{center}
		\centering
		\subfigure[]
		{
			\includegraphics[width=0.200\textwidth]{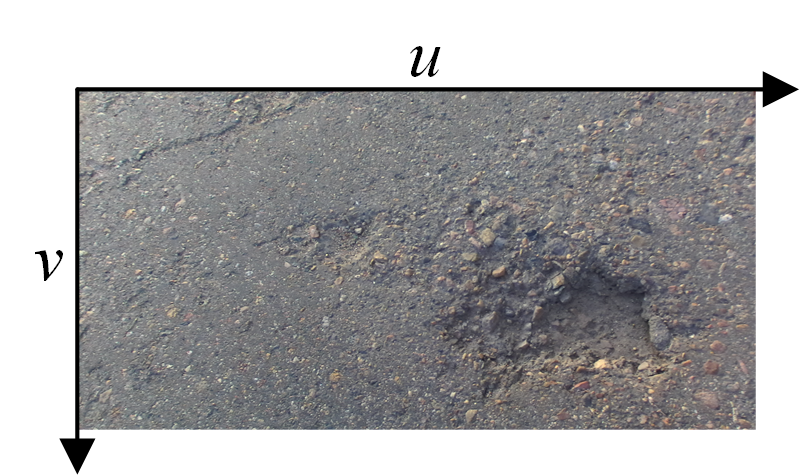}
			\label{fig.left}
		}
		\subfigure[]
		{
			\includegraphics[width=0.200\textwidth]{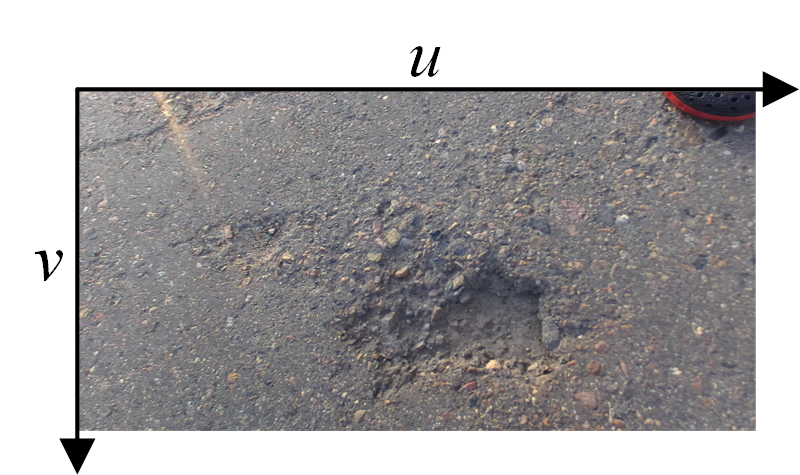}
			\label{fig.right}
		}\\
		\subfigure[]
		{
			\includegraphics[width=0.210\textwidth]{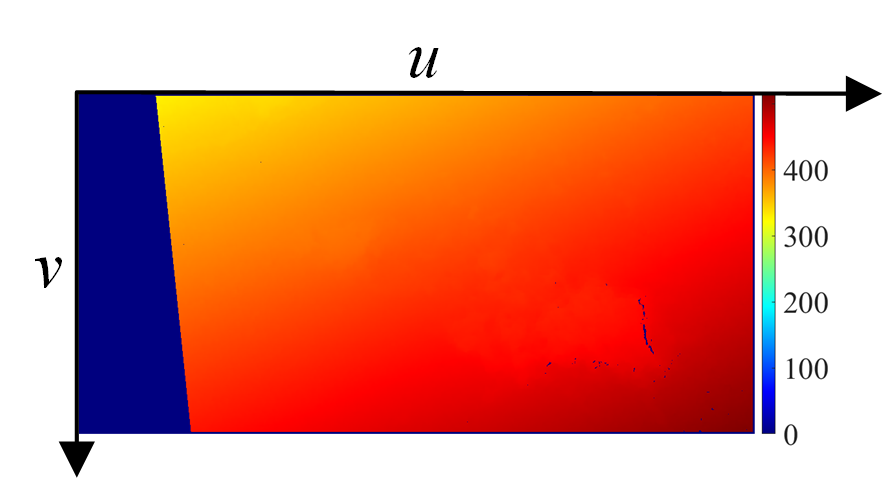}
			\label{fig.disp}
		}
		\subfigure[]
		{
			\includegraphics[width=0.0886\textwidth]{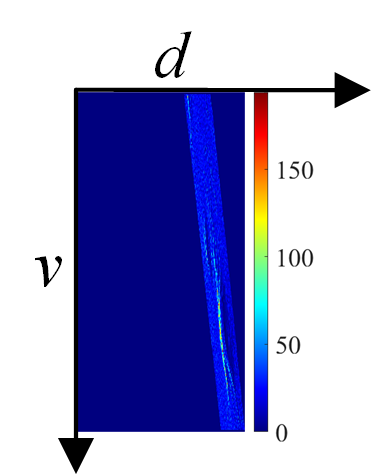}
			\label{fig.v_disp}
		}
		\caption{Disparity map when the roll angle does not equal zero; (a) left stereo image; (b) right stereo image; (c) dense disparity map; (d) v-disparity map.   }
		\label{fig.img_disp_v_disp}
	\end{center}
\end{figure}

However, in practice, the stereo rig baseline is not always perfectly parallel to the horizontal road surface. This fact can introduce a non-zero roll angle $\theta$, shown in  Fig. \ref{fig.roll_angle}, into the imaging process, where $T_c$ and $h$ represent the baseline and the mounting height of the stereo rig, respectively; $o_l^\mathscr{C}$ and $o_r^\mathscr{C}$ are the origins of the left and right camera coordinate systems, respectively. The stereo rig baseline can become perfectly parallel to the horizontal road surface by transforming the original world coordinate system $X^\mathscr{W}O^\mathscr{W}Y^\mathscr{W}$ to a new world coordinate system $X^{\mathscr{W}'}O^{\mathscr{W}}Y^{\mathscr{W}'}$. The original disparity map is shown in Fig. \ref{fig.img_disp_v_disp}, where readers can observe that the disparity values change gradually in the horizontal direction. This makes the way of representing the road disparity projections with a linear model problematic \cite{Fan2018b}. Therefore, the effects caused by the non-zero roll angle have to be eliminated before performing road region segmentation \cite{Fan2018b}. 

In recent years, many researchers have proposed to estimate the roll angle from the disparity map. For example,  in \cite{Ozgunalp2017} and \cite{Fan2018}, the road surface was assumed to be planar, and the disparities in a selected small area were modeled by a plane $g(u,v)=c_0+c_1u+c_2v$, where $u$ and $v$ are the horizontal and vertical disparity coordinates, respectively. The roll angle was then calculated as $\arctan(-c_1/c_2)$. However, finding a proper small area for plane modeling is always challenging, because such an area may contain an obstacle \cite{Fan2018b}. This can severely affect the plane modeling precision \cite{Evans2018}. In this regard, an iterative roll angle estimation algorithm was proposed in \cite{Evans2018}, where multiple small areas were selected for plane modeling. The roll angle was then estimated from  the optimum selected area. However, the aforementioned algorithms can only work for planar road surfaces. Therefore, in our previous work \cite{Fan2018b}, the projections of road disparities  were represented by a parabola, and the roll angle estimation was formulated as an energy minimization problem. To achieve greater processing efficiency, golden section search (GSS) algorithm \cite{Pedregal2006} was utilized to reduce the roll angle search range. However, this method still takes extensive iterations to converge to the minimum.  Therefore, the algorithm proposed in this paper solves the energy minimization problem using gradient descent (GD) \cite{Meza}, which greatly reduces the computational complexity of roll angle estimation.  

The remainder of this paper is structured as follows: Section \ref{sec.methodology} presents the proposed roll angle estimation algorithm. The experimental results for performance evaluation are illustrated in Section \ref{sec.experiments}. Finally, Section \ref{sec.conclusion&future_work} summarizes the paper and provides recommendations for future work. 

\section{Methodology}
\label{sec.methodology}
The input of our proposed algorithm is a dense disparity map, which was obtained from a stereo road image pair (see Fig. \ref{fig.left} and \ref{fig.right}) through our previously published disparity estimation algorithm \cite{Fan2018}. The estimated disparity map is shown in Fig. \ref{fig.disp}, and its v-disparity map is shown in Fig. \ref{fig.v_disp}. To rotate the disparity map around a given angle $\theta$, each set of original coordinates $[u,v]^\top$ is transformed to a set of new coordinates $[x(\theta),y(\theta)]^\top$ using the following equations:
\begin{equation}
x(\theta)=u\cos\theta+v\sin\theta,
\label{eq.x}
\end{equation}
\begin{equation}
y(\theta)=v\cos\theta-u\sin\theta.
\label{eq.y}
\end{equation}
The parabola representing the road disparity projections can, therefore, be written as follows:
\begin{equation}
\begin{split}
f(u,v,\theta)=\alpha_0+\alpha_1\cdot y(u,v,\theta)+\alpha_2\cdot y(u,v,\theta)^2.
\end{split}
\label{eq.f}
\end{equation}
The coefficients of $f(u,v,\theta)$ can, therefore, be estimated by solving an energy minimization problem as follows:
\begin{equation}
\hat{\boldsymbol{\alpha}}= \underset{\boldsymbol{\alpha}}{\arg\min}\ E(\theta)\ \ \text{s.t.} \ \theta\in(-\frac{\pi}{2},\frac{\pi}{2}],
\label{eq.energy_min_problem}
\end{equation}
where
\begin{equation}
\begin{split}
E(\theta)&=\sum_{i=0}^{n}\Big(d_i-f(u_i,v_i,\theta)  \Big)^2\\&=\big(\mathbf{d}-\mathbf{Y}(\theta)\boldsymbol{\alpha}(\theta)\big)^\top\big(\mathbf{d}-\mathbf{Y}(\theta)\boldsymbol{\alpha}(\theta)\big)
,
\end{split}
\label{eq.E}
\end{equation}
the column vector $\mathbf{d}=[d_0,d_1,\dots,d_n]^\top$ stores the disparity values, the column vector $\boldsymbol{\alpha}=[\alpha_0,\alpha_1,\alpha_2]^\top$ stores the coefficients of $f(u,v,\theta)$, and
\begin{equation}
\begin{split}
\mathbf{Y}(\theta)=
\begin{bmatrix}
1& y_0(\theta)& {y_0(\theta)}^2\\
1& y_1(\theta)& {y_1(\theta)}^2\\
\vdots& \vdots& \vdots\\
1& y_n(\theta)& {y_n(\theta)}^2\\
\end{bmatrix}.
\end{split}
\label{eq.Y}
\end{equation}
The energy minimization problem in (\ref{eq.energy_min_problem}) has a closed-form solution \cite{Weisstein2002}:
\begin{equation}
\boldsymbol{\alpha}=\mathbf{J}(\theta)\mathbf{d},
\label{eq.alpha}
\end{equation}
where 
\begin{equation}
\mathbf{J}(\theta)=\big(\mathbf{Y}(\theta)^\top\mathbf{Y}(\theta)\big)^{-1}\mathbf{Y}(\theta)^\top.
\label{eq.J}
\end{equation}
The minimum energy $E_\text{min}$ can also be obtained by combining (\ref{eq.E}) and (\ref{eq.alpha}): 
\begin{equation}
{E}_\text{min}(\theta)=\mathbf{d}^\top\mathbf{d}-\mathbf{d}^\top\mathbf{Y}(\theta)\mathbf{J}(\theta)\mathbf{d}.
\label{eq.E_min}
\end{equation}
According to \cite{Fan2018b}, the desirable roll angle corresponds to the minimum $E_\text{min}$. Therefore, the proposed roll angle estimation algorithm is equivalent to the following energy minimization problem:
\begin{equation}
	\hat{\theta}=\underset{\theta}{\arg\min}\ E_\text{min}(\theta) \ \ \text{s.t.} \ \theta\in(-\frac{\pi}{2},\frac{\pi}{2}].
	\label{eq.e_min_min}
\end{equation}
However, deriving the numeric solution of (\ref{eq.e_min_min}) is significantly challenging  \cite{Fan2018b}. Therefore, the proposed algorithm utilizes GD to approximate $\theta$ by formulating (\ref{eq.e_min_min}) as an iterative optimization problem as follows \cite{rao2009engineeringoptimizationtheory}:
\begin{equation}
\theta^{(k+1)}=\theta^{(k)}-\lambda\nabla{E}_\text{min}({\theta}^{(k)}),\ \ \ k\in\mathbb{N}^0,
\label{eq.e_min_gd}
\end{equation}
where 
\begin{equation}
\nabla E_\text{min}(\theta)=-2\mathbf{d}^\top\mathbf{M}(\theta)\mathbf{d}, 
\label{eq.d_e_min}
\end{equation}
\begin{equation}
\begin{split}
\mathbf{M}(\theta)=\Big[\mathbf{I}-\mathbf{Y}(\theta)\mathbf{J}(\theta)\Big]\nabla\mathbf{Y}(\theta)\mathbf{J}(\theta),
\end{split}
\label{eq.M}
\end{equation}
$\mathbf{I}$ is an identity matrix, and $\lambda$ is the learning rate. If $\lambda$ is too high, (\ref{eq.e_min_gd}) may overshoot the minimum. On the other hand, if $\lambda$ is set to a relatively low value, the convergence of (\ref{eq.e_min_gd}) may take too many iterations \cite{Pedregal2006}. Therefore, selecting a proper $\lambda$ is always essential for  GD. Instead of fixing the learning rate with a constant, we utilize backtracking line search to produce an adaptive learning rate: 
\begin{equation}
\lambda^{(k+1)}=\frac{\lambda^{(k)}\nabla{E}_\text{min}({\theta}^{(k)})}{\nabla{E}_\text{min}({\theta}^{(k)})-\nabla{E}_\text{min}({\theta}^{(k+1)})},\ \ \  k\in\mathbb{N}^0.
\label{eq.adaptive_lambda}
\end{equation}
The initial learning rate $\lambda^{(0)}$ is set to 26. We will discuss the selection of $\lambda^{(0)}$ in Section \ref{sec.experiments}. 
The initial approximation $\theta^{(0)}$ is set to $0$, because in practical experiments the absolute value of the roll angle is usually not that high. The optimization iterates until the absolute difference between $\theta^{(k)}$ and $\theta^{(k+1)}$ is smaller than a preset threshold $\delta_\theta$.  The performance of the proposed roll angle estimation algorithm will be discussed in Section \ref{sec.experiments}. 
\begin{figure*}[!t]
	\begin{center}
		\centering
		\includegraphics[width=0.90\textwidth]{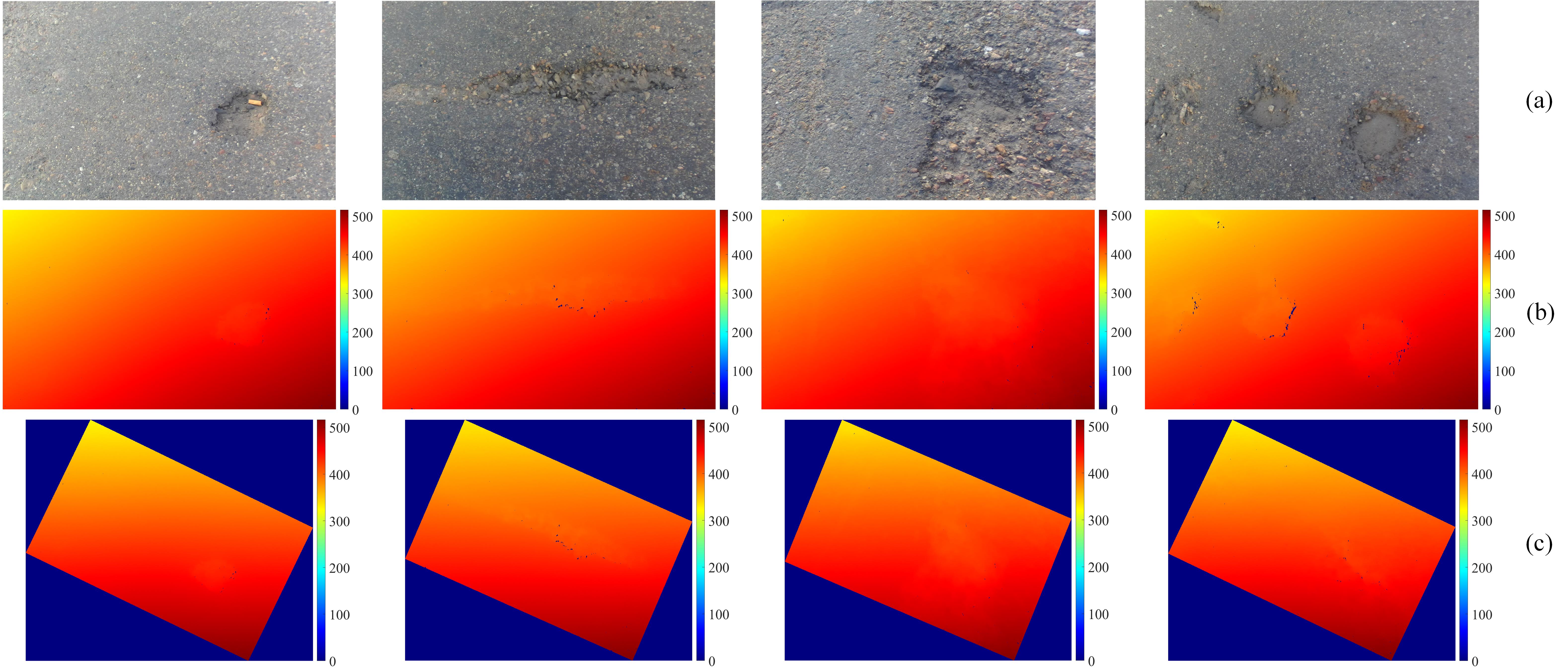}
		\caption{Examples of the roll angle estimation experimental results; (a) left images; (b) original disparity maps; (c) disparity maps rotated around the estimated roll angles.   }
		\label{fig.experimental_results}
	\end{center}
\end{figure*}
\section{Experimental Results}
\label{sec.experiments}
In this section, we evaluate the performance of the proposed roll angle estimation algorithm both qualitatively and quantitatively. The proposed algorithm was implemented in Matlab 2019a on an Intel Core i7-8700K CPU (3.7 GHz) using a single thread.

\begin{figure}[!t]
	\begin{center}
		\centering
		\includegraphics[width=0.48\textwidth]{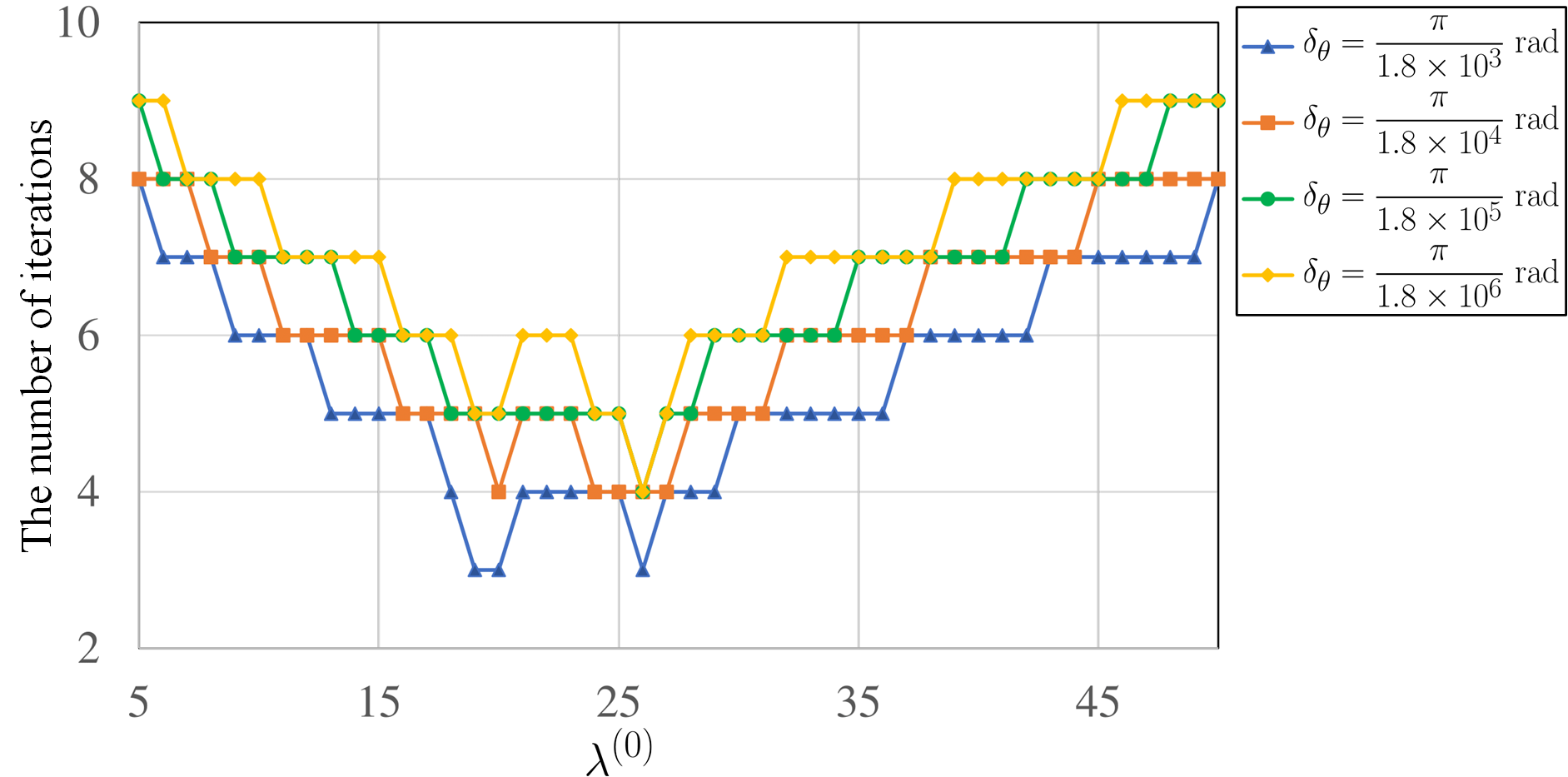}
		\caption{The number of iterations with respect to different $\lambda^{(0)}$ and $\delta_\theta$.   }
		\label{fig.params_selection}
	\end{center}
\end{figure}
Firstly, we discuss the selection of $\lambda^{(0)}$. In our experiments, we utilized a ZED  stereo camera\footnote{https://www.stereolabs.com/} to capture stereo road image pairs. The disparity maps estimated from the stereo image pairs were then utilized to estimate the roll angles. Some experimental results of our pothole datasets\footnote{www.ruirangerfan.com} are shown in Fig. \ref{fig.experimental_results}, where readers can observe that the disparities on each row in the rotated disparity maps have similar values. Therefore, the proposed algorithm can effectively estimate the roll angle from a given disparity map. 

Furthermore, we select  a range of $\lambda^{(0)}$ and record the number of iterations that GD takes to converge with respect to different $\delta_\theta$, as shown in Fig. \ref{fig.params_selection}, where readers can see that $\lambda^{(0)}=26$ is the optimum value for different $\delta_\theta$. When $\delta_\theta=\frac{\pi}{1.8\times10^3}$ rad ($0.1^\circ$), GD takes three iterations to converge, while GD iterates four times, when $\delta_\theta$ is set to $\frac{\pi}{1.8\times10^4}$ rad ($0.01^\circ$), $\frac{\pi}{1.8\times10^5}$ rad ($0.001^\circ$) or $\frac{\pi}{1.8\times10^6}$ rad ($0.0001^\circ$).

\begin{figure*}[!t]
	\begin{center}
		\centering
		\includegraphics[width=0.80\textwidth]{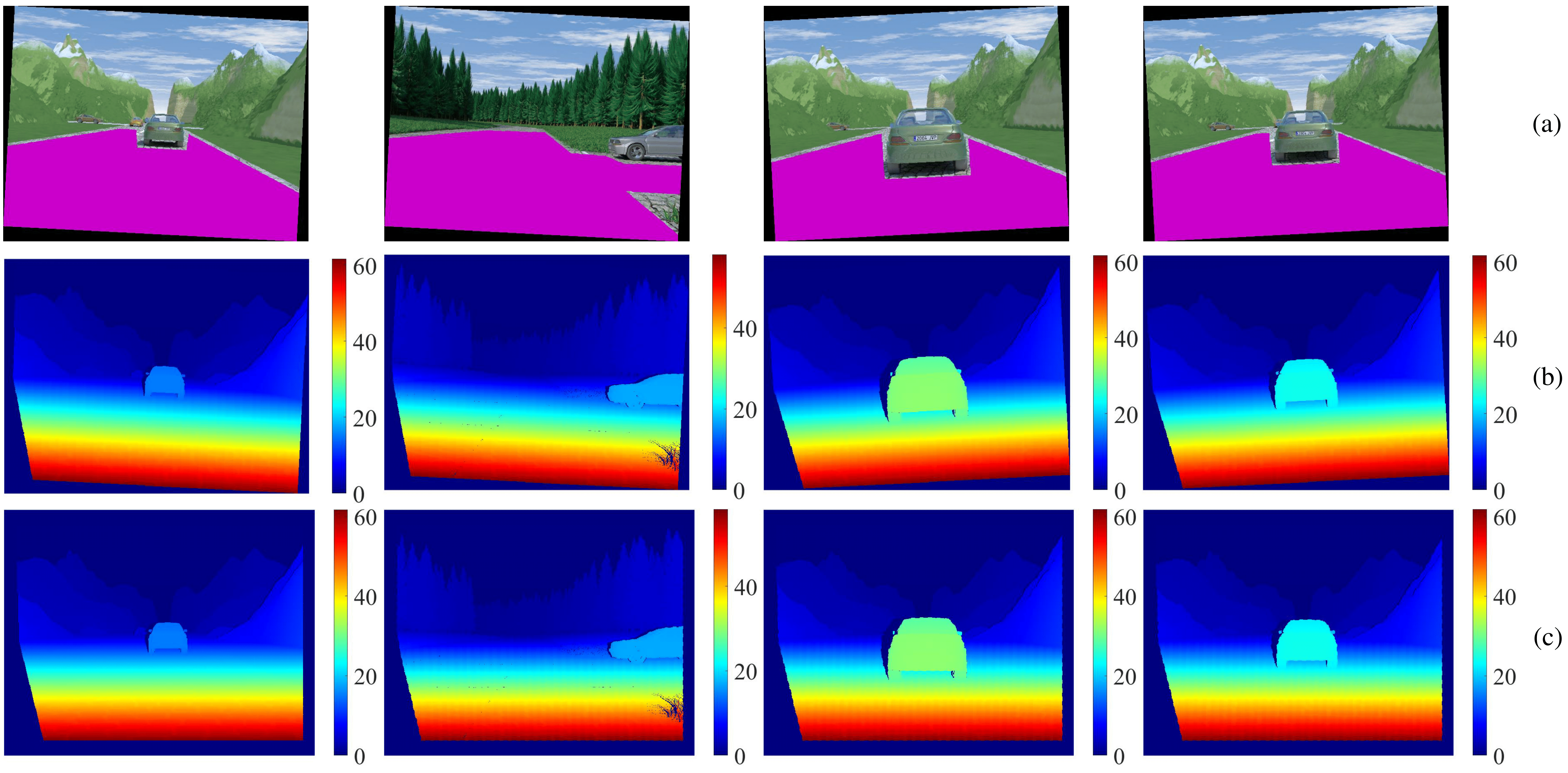}
		\caption{Experimental results of the EISATS stereo datasets; (a) left stereo images, the areas in magenta are our manually selected road regions; (b) disparity maps; (c) disparity maps rotated around the estimated roll angles.   }
		\label{fig.eisats_results}
	\end{center}
\end{figure*}

Moreover, we compare the computational efficiency of the proposed algorithm with \cite{Fan2018b}. As discussed in \cite{Fan2018b}, the computational complexity of GSS is $\mathcal{O}(\log_{k}{\frac{\delta_\theta}{\pi}})$, where $k=0.618$ is the golden section ratio\cite{Pedregal2006}. 
The comparison of the computational efficiency between GSS and GD is shown in Table \ref{table.gss_gd_comparison}, where we can see that GSS takes more iterations to converge with the decrease of $\delta_\theta$, while the performance of GD largely remains the same. Therefore, the proposed roll angle estimation algorithm performs more robustly  than \cite{Fan2018b} with respect to different thresholds $\delta_\theta$. 
%\begin{table}[!h]
%	\begin{center}
%		\footnotesize
%		\caption{Comparison between GSS and GD in terms of the number of iterations. $\lambda^{(0)}=26$. }
%		\label{table.gss_gd_comparison}
%		\begin{tabular}{|c|c|c|c|c|c|}
%			\hline
%			$\delta_\theta$ (rad) & GSS & GD  \\
%			\hline
%			$\frac{\pi}{1.8\times10^3}$ & 16 & 3  \\
%			$\frac{\pi}{1.8\times10^4}$ & 21 & 4	\\
%			$\frac{\pi}{1.8\times10^5}$ & 26 & 4 \\
%			$\frac{\pi}{1.8\times10^6}$ & 30  & 4\\
%			\hline
%		\end{tabular}
%	\end{center}
%\end{table}
\begin{table}[!t]
	\begin{center}
		\vspace{0in}
		\footnotesize
		\caption{Comparison between GSS and GD in terms of the number of iterations. $\lambda^{(0)}=26$.}
		\label{table.gss_gd_comparison}
		\begin{tabular}{|c|c|c|c|c|}
			\hline
			\multirow{2}{*}{Method} & \multicolumn{4}{c|}{$\delta_\theta$ (rad)} \\
			\cline{2-5}
			& $\frac{\pi}{1.8\times10^3}$ & $\frac{\pi}{1.8\times10^4}$ & $\frac{\pi}{1.8\times10^5}$ & $\frac{\pi}{1.8\times10^6}$ \\
			\hline
			GSS & 16 & 21 & 26 & 30\\
			\hline
			GD & 3 & 4 & 4 & 4\\
			\hline
		\end{tabular}
	\end{center}
\end{table}

However, GD involves the computations of both $\nabla E_\text{min}$ and $\lambda$, which are more complex than the computation of $E$. Therefore, we also compare the runtime of GSS and GD with respect to different thresholds $\delta_\theta$, as shown in Table \ref{table.gss_gd_runtime}. Readers can observe that the runtime of GSS goes up gradually with the decrease of $\delta_\theta$, while the runtime of GD changes slightly with respect to different $\delta_\theta$. Overall, the proposed algorithm performs much faster than \cite{Fan2018b}.
\begin{table}[!t]
	\begin{center}
		\vspace{0in}
		\footnotesize
		\caption{Comparison between GSS and GD in terms of runtime (in seconds). $\lambda^{(0)}=26$.}
		\label{table.gss_gd_runtime}
		\begin{tabular}{|c|c|c|c|c|}
			\hline
			\multirow{2}{*}{Method} & \multicolumn{4}{c|}{$\delta_\theta$ (rad)} \\
			\cline{2-5}
			& $\frac{\pi}{1.8\times10^3}$ & $\frac{\pi}{1.8\times10^4}$ & $\frac{\pi}{1.8\times10^5}$ & $\frac{\pi}{1.8\times10^6}$ \\
			\hline
			GSS & 1.551 & 1.986 & 2.457 & 2.810\\
			\hline
			GD & 0.412 & 0.401 & 0.511 & 0.512\\
			\hline
		\end{tabular}
	\end{center}
\end{table}

%\begin{table}[!h]
%	\begin{center}
%		\footnotesize
%		\caption{Comparison between GSS and GD in terms of runtime (in seconds). $\lambda^{(0)}=26$. }
%		\label{table.gss_gd_runtime}
%		\begin{tabular}{|c|c|c|c|c|c|}
%			\hline
%			$\delta_\theta$ (rad) & GSS & GD  \\
%			\hline
%			$\frac{\pi}{1.8\times10^3}$ & 1.551 & 0.412  \\
%			$\frac{\pi}{1.8\times10^4}$ & 1.986 & 0.401	\\
%			$\frac{\pi}{1.8\times10^5}$ & 2.457 & 0.511 \\
%			$\frac{\pi}{1.8\times10^6}$ & 2.810  & 0.512\\
%			\hline
%		\end{tabular}
%	\end{center}
%\end{table}

\begin{figure}[!t]
	\begin{center}
		\centering
		\includegraphics[width=0.44\textwidth]{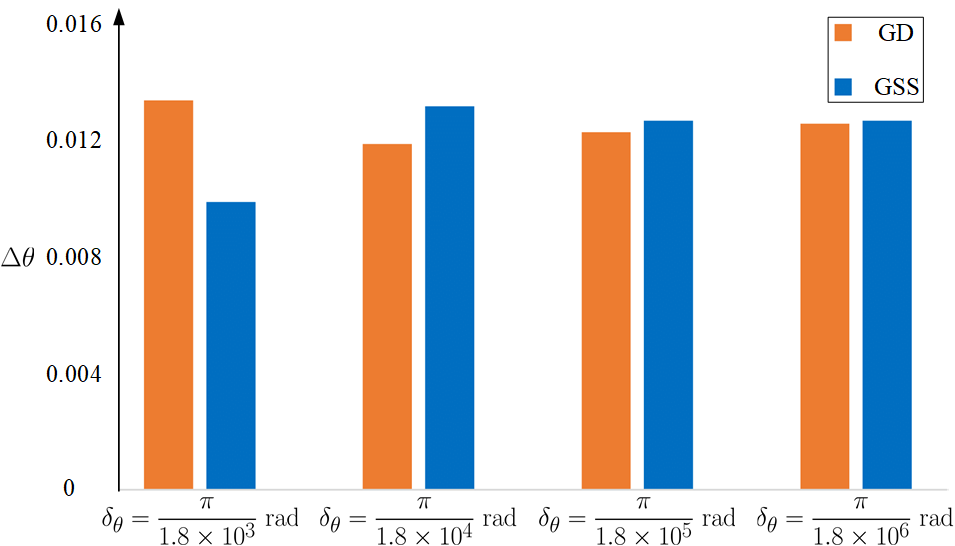}
		\caption{Comparison between GD and GSS in terms of the roll angle accuracy.   }
		\label{fig.accuracy_evaluation}
	\end{center}
\end{figure}

To quantify the accuracy of the proposed roll angle estimation algorithm, we utilize a synthesized stereo dataset from EISATS\footnote{https://ccv.wordpress.fos.auckland.ac.nz/} \cite{Vaudrey2008, Wedel2008}, where the roll angle is perfectly zero. We manually rotate the disparity maps around a given angle, and then estimate the roll angles from the rotated disparity maps. Examples of the EISATS stereo datasets are shown in Fig. \ref{fig.eisats_results}, where readers can see that the effects caused by the roll angle are successfully eliminated by performing the proposed algorithm. Furthermore, we compare the accuracy of the estimated roll angles between the proposed method and \cite{Fan2018b} with respect to different $\delta_\theta$. The comparison results are shown in Fig. \ref{fig.accuracy_evaluation}, where $\Delta\theta$ represents the average difference between the actual and  estimated roll angles. From Fig. \ref{fig.accuracy_evaluation}, we can find that GD performs more accurately than GSS when $\delta_\theta$ is set to a lower value. Additionally, the computational efficiency of GD is also greater than that of GSS. Therefore, GD is a better solution than GSS in terms of solving the energy minimization problem in roll angle estimation.

\section{Conclusion and Future Work}
\label{sec.conclusion&future_work}
In this paper, a robust GD-based algorithm was proposed to estimate the roll angle from disparity maps.  The proposed algorithm was developed based on our previously published work, where roll angle estimation was formulated as an energy minimization problem, which was then solved using GSS. The experimental results demonstrated that the proposed algorithm outperforms our previous method in terms of both accuracy and efficiency, and that it can successfully eliminate the effects caused by the non-zero roll angle.  In the future, we aim to design a neural network to estimate the roll angle with an unsupervised learning strategy.

\section*{Acknowledgment}
This work is supported by  grants from  the Research Grants Council of the Hong Kong SAR Government, China (No. 11210017, No. 16212815, No. 21202816, NSFC U1713211) awarded to Prof. Ming Liu. This work is also supported by grants from the Shenzhen Science, Technology and Innovation Commission (JCYJ20170818153518789), National Natural Science Foundation of China (No. 61603376) and Guangdong Innovation and Technology Fund (No. 2018B050502009) awarded to Dr. Lujia Wang. This work is also supported by funding awarded to Prof. Ioannis Pitas received from the European Union Horizon 2020 research and innovation programme, under grant agreement No. 731667 (Multidrone). 

\clearpage
\balance
\bibliographystyle{IEEEtran}

\begin{thebibliography}{10}
	\providecommand{\url}[1]{#1}
	\csname url@samestyle\endcsname
	\providecommand{\newblock}{\relax}
	\providecommand{\bibinfo}[2]{#2}
	\providecommand{\BIBentrySTDinterwordspacing}{\spaceskip=0pt\relax}
	\providecommand{\BIBentryALTinterwordstretchfactor}{4}
	\providecommand{\BIBentryALTinterwordspacing}{\spaceskip=\fontdimen2\font plus
		\BIBentryALTinterwordstretchfactor\fontdimen3\font minus
		\fontdimen4\font\relax}
	\providecommand{\BIBforeignlanguage}[2]{{%
			\expandafter\ifx\csname l@#1\endcsname\relax
			\typeout{** WARNING: IEEEtran.bst: No hyphenation pattern has been}%
			\typeout{** loaded for the language `#1'. Using the pattern for}%
			\typeout{** the default language instead.}%
			\else
			\language=\csname l@#1\endcsname
			\fi
			#2}}
	\providecommand{\BIBdecl}{\relax}
	\BIBdecl
	
	\bibitem{Fan2018e}
	R.~Fan, ``Real-time computer stereo vision for automotive applications,'' Ph.D.
	dissertation, University of Bristol, 2018.
	
	\bibitem{Trucco1998}
	E.~Trucco and A.~Verri, \emph{Introductory techniques for 3-D computer
		vision}.\hskip 1em plus 0.5em minus 0.4em\relax Prentice Hall Englewood
	Cliffs, 1998, vol. 201.
	
	\bibitem{Fan2018f}
	R.~Fan, Y.~Liu, M.~J. Bocus, and M.~Liu, ``Real-time subpixel fast bilateral
	stereo,'' \emph{arXiv preprint arXiv:1807.02044}, 2018.
	
	\bibitem{Qian1997}
	N.~Qian, ``Binocular disparity and the perception of depth,'' \emph{Neuron},
	vol.~18, no.~3, pp. 359--368, 1997.
	
	\bibitem{Fan2017a}
	R.~Fan and N.~Dahnoun, ``Real-time implementation of stereo vision based on
	optimised normalised cross-correlation and propagated search range on a
	gpu,'' in \emph{2017 IEEE International Conference on Imaging Systems and
		Techniques (IST)}.\hskip 1em plus 0.5em minus 0.4em\relax IEEE, 2017, pp.
	1--6.
	
	\bibitem{Fan2018g}
	R.~Fan, Y.~Liu, X.~Yang, M.~J. Bocus, N.~Dahnoun, and S.~Tancock, ``Real-time
	stereo vision for road surface 3-d reconstruction,'' in \emph{2018 IEEE
		International Conference on Imaging Systems and Techniques (IST)}.\hskip 1em
	plus 0.5em minus 0.4em\relax IEEE, 2018, pp. 1--6.
	
	\bibitem{Fan2019}
	R.~Fan, J.~Jiao, J.~Pan, H.~Huang, S.~Shen, and M.~Liu, ``Real-time dense
	stereo embedded in a uav for road inspection,'' \emph{arXiv preprint
		arXiv:1904.06017}.
	
	\bibitem{Labayrade2002}
	R.~Labayrade, D.~Aubert, and J.-P. Tarel, ``Real time obstacle detection in
	stereovision on non flat road geometry through" v-disparity"
	representation,'' in \emph{Intelligent Vehicle Symposium, 2002. IEEE},
	vol.~2.\hskip 1em plus 0.5em minus 0.4em\relax IEEE, 2002, pp. 646--651.
	
	\bibitem{Soquet2007}
	N.~Soquet, D.~Aubert, and N.~Hautiere, ``Road segmentation supervised by an
	extended v-disparity algorithm for autonomous navigation,'' in \emph{2007
		IEEE Intelligent Vehicles Symposium}.\hskip 1em plus 0.5em minus 0.4em\relax
	IEEE, 2007, pp. 160--165.
	
	\bibitem{Fan2018d}
	R.~Fan and N.~Dahnoun, ``Real-time stereo vision-based lane detection system,''
	\emph{Measurement Science and Technology}, vol.~29, no.~7, p. 074005, 2018.
	
	\bibitem{Yiruo2013}
	D.~Yiruo, W.~Wenjia, and K.~Yukihiro, ``Complex ground plane detection based on
	v-disparity map in off-road environment,'' in \emph{2013 IEEE Intelligent
		Vehicles Symposium (IV)}.\hskip 1em plus 0.5em minus 0.4em\relax IEEE, 2013,
	pp. 1137--1142.
	
	\bibitem{Broggi2005}
	A.~{Broggi}, C.~{Caraffi}, R.~I. {Fedriga}, and P.~{Grisleri}, ``Obstacle
	detection with stereo vision for off-road vehicle navigation,'' in
	\emph{Proc. IEEE Computer Society Conf. Computer Vision and Pattern
		Recognition (CVPR'05) - Workshops}, Sep. 2005, p.~65.
	
	\bibitem{hu2005complete}
	Z.~Hu, F.~Lamosa, and K.~Uchimura, ``A complete uv-disparity study for
	stereovision based 3d driving environment analysis,'' in \emph{3-D Digital
		Imaging and Modeling, 2005. 3DIM 2005. Fifth International Conference on},
	2005, pp. 204--211.
	
	\bibitem{Chen2018}
	T.~{Chen}, B.~{Chen}, X.~{Zhang}, and Z.~{Yuan}, ``Free space detection using
	stereo confidence metrics and obstacle position probability maps,'' in
	\emph{Proc. 14th IEEE Int. Conf. Signal Processing (ICSP)}, Aug. 2018, pp.
	1071--1075.
	
	\bibitem{Oniga2015}
	F.~{Oniga}, E.~{Sarkozi}, and S.~{Nedevschi}, ``Fast obstacle detection using
	u-disparity maps with stereo vision,'' in \emph{Proc. IEEE Int. Conf.
		Intelligent Computer Communication and Processing (ICCP)}, Sep. 2015, pp.
	203--207.
	
	\bibitem{Song2017}
	W.~{Song}, M.~{Fu}, Y.~{Yang}, M.~{Wang}, X.~{Wang}, and A.~{Kornhauser},
	``Real-time lane detection and forward collision warning system based on
	stereo vision,'' in \emph{Proc. IEEE Intelligent Vehicles Symp. (IV)}, Jun.
	2017, pp. 493--498.
	
	\bibitem{Ma2018}
	H.~Ma, Y.~Ma, J.~Jiao, M.~U.~M. Bhutta, M.~J. Bocus, L.~Wang, M.~Liu, and
	R.~Fan, ``Multiple lane detection algorithm based on optimised dense
	disparity map estimation,'' in \emph{2018 IEEE International Conference on
		Imaging Systems and Techniques (IST)}.\hskip 1em plus 0.5em minus 0.4em\relax
	IEEE, 2018, pp. 1--5.
	
	\bibitem{Fan2018b}
	R.~Fan, M.~J. Bocus, and N.~Dahnoun, ``A novel disparity transformation
	algorithm for road segmentation,'' \emph{Information Processing Letters},
	vol. 140, pp. 18--24, 2018.
	
	\bibitem{Ozgunalp2017}
	U.~Ozgunalp, R.~Fan, X.~Ai, and N.~Dahnoun, ``Multiple lane detection algorithm
	based on novel dense vanishing point estimation,'' \emph{IEEE Transactions on
		Intelligent Transportation Systems}, vol.~18, no.~3, pp. 621--632, 2017.
	
	\bibitem{Fan2018}
	R.~Fan, X.~Ai, and N.~Dahnoun, ``Road surface {3D} reconstruction based on
	dense subpixel disparity map estimation,'' \emph{IEEE Transactions on Image
		Processing}, vol.~PP, no.~99, p.~1, 2018.
	
	\bibitem{Evans2018}
	M.~Evans, R.~Fan, and N.~Dahnoun, ``Iterative roll angle estimation from dense
	disparity map,'' in \emph{2018 7th Mediterranean Conference on Embedded
		Computing (MECO)}.\hskip 1em plus 0.5em minus 0.4em\relax IEEE, 2018.
	
	\bibitem{Pedregal2006}
	P.~Pedregal, \emph{Introduction to optimization}.\hskip 1em plus 0.5em minus
	0.4em\relax Springer Science \& Business Media, 2006, vol.~46.
	
	\bibitem{Meza}
	J.~C. Meza, S.~Profile, J.~C. Meza, R.~A. Oliva, E.~O.~L. Berkeley, P.~D.
	Hough, and P.~J. Williams, ``Nonlinear optimization.''
	
	\bibitem{Weisstein2002}
	E.~W. Weisstein, ``Least squares fitting,'' 2002.
	
	\bibitem{rao2009engineeringoptimizationtheory}
	S.~S. Rao and S.~S. Rao, \emph{Engineering Optimization: Theory and
		Practice}.\hskip 1em plus 0.5em minus 0.4em\relax Wiley, 2009.
	
	\bibitem{Vaudrey2008}
	T.~Vaudrey, C.~Rabe, R.~Klette, and J.~Milburn, ``Differences between stereo
	and motion behaviour on synthetic and real-world stereo sequences,'' in
	\emph{Image and Vision Computing New Zealand, 2008. IVCNZ 2008. 23rd
		International Conference}.\hskip 1em plus 0.5em minus 0.4em\relax IEEE, 2008,
	pp. 1--6.
	
	\bibitem{Wedel2008}
	A.~Wedel, C.~Rabe, T.~Vaudrey, T.~Brox, U.~Franke, and D.~Cremers, ``Efficient
	dense scene flow from sparse or dense stereo data,'' in \emph{European
		conference on computer vision}.\hskip 1em plus 0.5em minus 0.4em\relax
	Springer, 2008, pp. 739--751.
	
\end{thebibliography}

\end{document}